\documentclass[runningheads]{llncs}
\usepackage{graphicx}
\usepackage{amsmath}
\usepackage{amssymb}
\usepackage{makecell}
\usepackage{stackengine}
\begin{document}
\title{Exploring Temporal Differences in 3D Convolutional Neural Networks}
%
%
\author{Gagan Kanojia
 \and
Sudhakar Kumawat\inst{} \and
Shanmuganathan Raman\inst{}}
%
%
\institute{IIT Gandhinagar}
%
\maketitle              
\begin{abstract}
\noindent Traditional 3D convolutions are computationally expensive, memory intensive, and due to large number of parameters, they often tend to overfit. On the other hand, 2D CNNs are less computationally expensive and less memory intensive than 3D CNNs and have shown remarkable results in applications like image classification and object recognition. However, in previous works, it has been observed that they are inferior to 3D CNNs when applied on a spatio-temporal input. In this work, we propose a convolutional block which extracts the spatial information by performing a 2D convolution and extracts the temporal information by exploiting temporal differences, i.e., the change in the spatial information at different time instances, using simple operations of shift, subtract and add without utilizing any trainable parameters. The proposed convolutional block has same number of parameters as of a 2D convolution kernel of size $n\times n$, i.e. $n^2$, and has $n$ times lesser parameters than an $n\times n \times n$ 3D convolution kernel. We show that the 3D CNNs perform better when the 3D convolution kernels are replaced by the proposed convolutional blocks. We evaluate the proposed convolutional block on UCF101 and ModelNet datasets. All the codes and pretrained models will be publicly available at \_.

\end{abstract}
\section{Introduction}
\label{sec:intro}
Lately, 3D convolutional neural networks are gaining popularity over the 2D CNNs when the task is to deal with 3D data representations which could be videos, shapes or other formats \cite{hara2018can,tran2017convnet}. This is because 2D CNN lack in exploiting the temporal information. 3D CNNs are more proficient than 2D CNNs in extracting temporal information and utilizing it to perform specific tasks. It has been shown that a 3D CNN of same depth as that of a 2D CNN performs better on tasks like action recognition \cite{hara2018can,tran2018closer}. However, this proficiency comes with a cost in terms of the number of learnable parameters, memory requirements, and risks of overfitting. For example, 3D ResNet (18 layers) \cite{hara2018can} has around 3 times more parameters than the 2D ResNet (18 layers) \cite{he2016deep}.\\
In this work, our focus is on acquiring both spatial and temporal structure of the 3D data while reducing the cost in terms of trainable parameters. We propose a convolutional block which exploits both the spatial information and the temporal information by utilizing a 2D convolution and temporal differences, i.e., the change in the spatial information at different time instances, using simple operations of shift, subtract and add. We have also incorporated temporal max pooling in order to downsample the temporal depth of the feature maps along the depth of the network. None of the operations other than 2D convolution require trainable parameters which makes the number of trainable parameters of the proposed convolutional block equal to the 2D convolution kernel with same kernel size. The major contributions of the work are as follows. \textbf{(a)} We propose a novel convolutional block which captures spatial information by performing a 2D convolution and captures temporal information using simple operations of shift, subtract and add. \textbf{(b)} We reduce the number of parameters by $n$ times by replacing the 3D convolution kernel of size $n\times n\times n$ with the proposed convolution block comprising a 2D convolution kernel of size $1\times n\times n$. \textbf{(c)} We show that the proposed convolutional block helps the 3D CNNs to perform better while utilizing lesser parameters than the 3D convolution kernels.
\section{Related work}
\label{sec:related}
In recent years, 2D CNNs have been dominating several applications of computer vision like object detection \cite{he2016deep} and image classification\cite{he2016deep}. However, they lack in extracting the temporal information present in the spatio-temporal data \cite{tran2018closer}. There are works which extend the 2D CNNs on videos by processing the video frames individually and then combining the extracted information along the temporal dimension to obtain the output \cite{xu2015discriminative,girdhar2017actionvlad}.
Recently, 3D CNNs have shown great potential in dealing with the spatio-temporal data or 3D CAD models as inputs \cite{tran2015learning,zhi2017lightnet,maturana2015voxnet}. It has been observed that 3D CNNs are much better in exploiting the temporal information than 2D CNNs\cite{tran2018closer}. However, 3D CNNs are computationally expensive and they are prone to overfit due to their large number of parameters. Hence, the researchers moved on to find better and more efficient ways of mimicking 3D convolutions. There has been notable advances in the separable convolutions in 2D CNNs to reduce the space-time complexity \cite{sandler2018mobilenetv2,chollet2017xception,xie2017aggregated}. In many works, the idea of separable convolutions has been extended to 3D CNNs \cite{sun2015human,xie2018rethinking,qiu2017learning,tran2018closer}. In \cite{qiu2017learning}, the authors proposed the idea of replacing the 3D convolution kernel by a 2D convolution kernel to capture the spatial information followed by a 1D convolution kernel to convolve along the temporal direction. They showed that the proposed technique has several advantages, like parameter reduction and better performance, over the 3D convolutions, which has been further explored in \cite{tran2018closer}. Temporal differences has been explored in few recent works \cite{wang2016temporal,lee2018motion}. Wang \emph{et al.} \cite{wang2016temporal} use difference in two frames as the approximation of motion information. Similarly, Lee \emph{et al.} \cite{lee2018motion} propose a motion block which extracts features using spatial and temporal shifts. In this work, we only rely on the temporal differences. Instead of relying on only the adjacent frames, we compute aggregated temporal differences over several frames. The proposed SSA Layer does not involve any trainable parameter to extract temporal information via temporal differences. Our focus is to propose an efficient alternative to the 3D convolution filters which utilizes lesser parameters without compromising the performance.
\begin{figure}[t]
	\centering
	\includegraphics[width = 0.85\linewidth]{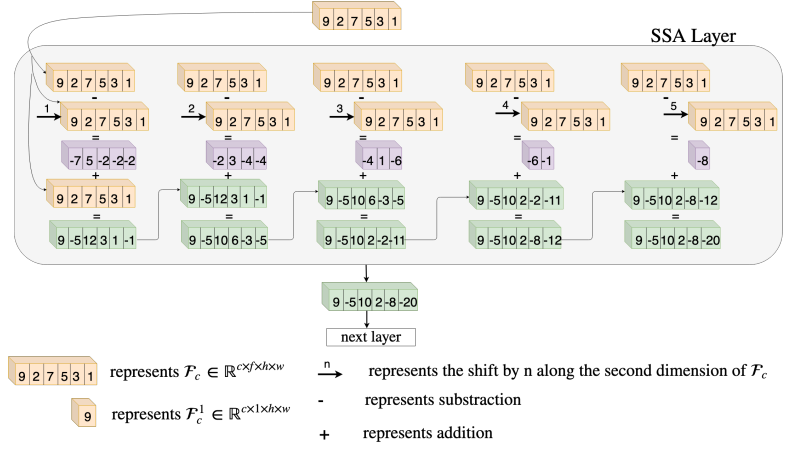}
	\caption{An illustration of SSA Layer}
	\label{fig:ssa_layer}
\end{figure}
\section{Proposed Approach}
\label{sec:conv_block}
In this section, we discuss the proposed convolutional block which extracts both spatial and temporal information. The proposed convolutional block has three parts: 2D convolution kernel, SSA layer, and temporal pooling layer. Here, SSA stands for Shift, Subtract and Add. Let the input to the proposed convolutional block be $\mathcal{X} \in \mathbb{R}^{c\times f \times h \times w}$. Here, $\mathcal{X}$ is the output feature maps of the previous convolutional block or layer, $c$ is number of channels, $f$ corresponds to the temporal depth, and $h$ and $w$ are the height and width of $\mathcal{X}$, respectively.\\
\textbf{2D convolution.} In traditional 3D CNNs, the feature maps are convolved with a 3D filter $\hat{g} \in \mathbb{R}^{c\times k \times k \times k}$ with $c$ channels and kernel size $k \times k \times k$ \cite{hara2018can}. In the proposed framework, first we obtain $\mathcal{X}_c = \mathcal{X}\star g$.
Here, $\star$ stands for convolution, and $g$ is a 2D filter of kernel size $1\times k \times k$ and $c$ channels. The purpose of the 2D convolution is to extract the spatial information present in the input feature maps\cite{zeiler2014visualizing}. We, then, pass $\mathcal{X}_c$ through the proposed SSA layer to obtain the temporal structure of the feature maps.\\
\noindent\textbf{SSA Layer.} SSA stands for Shift, Subtract and Add operations performed in the SSA layer. The purpose of the SSA layer is to extract the temporal information present in the spatio-temporal data. For example, in action recognition, motion features extracted from the videos can hold important information. In order to capture the motion information, optical flow techniques can be used \cite{dosovitskiy2015flownet}. However, capturing optical flow is in itself a computationally expensive task which can require a dedicated network \cite{dosovitskiy2015flownet}. In the proposed SSA layer, we rely on temporal differences, i.e., the change in the spatial information at different time instances, to extract the necessary temporal information present in the spatio-temporal data.In the case of action recognition, temporal differences can provide the rough extimate of the location of moving objects or non-rigid bodies \cite{park2013exploring}.  However, there is a possibility that there has not been enough change occurred in the adjacent frames. Hence, we take multiple frames into the consideration.  The difference could be due to motion like in the case of action recognition or due to the structure of the input, like in the case of shapes. This makes the SSA layer to be used in a more general sense.\\
Let the input to the SSA layer be $\mathcal{X}_c \in \mathbb{R}^{c\times f \times h \times w}$. Here, $c$ is the number of channels, $f$ is the temporal depth, and $h$ and $w$ are the height and width of $\mathcal{X}_c$, respectively. We obtain the temporal differences between the volumes of the feature map $\mathcal{X}_c$ along the temporal depth. Let $\{\mathcal{X}_c^1,\mathcal{X}_c^2, \mathcal{X}_c^3,\ldots,\mathcal{X}_c^f\} \in \mathbb{R}^{c\times 1 \times h \times w}$ be the volumes of the feature map $\mathcal{X}_c$ along the temporal depth. $\mathcal{X}_c$ is passed through the SSA layer to obtain $\mathcal{X}_s \in \mathbb{R}^{c\times f \times h \times w} $ as shown in Eq. \ref{eq:ssa_layer2}. Here, $\mathcal{X}_s$ is obtained by concatenating $\{\mathcal{X}_s^1,\mathcal{X}_s^2, \mathcal{X}_s^3,\ldots,\mathcal{X}_s^f \}\in \mathbb{R}^{c\times 1 \times h \times w}$ along the temporal dimension.
\begin{equation}
\mathcal{X}_s^i = \mathcal{X}_c^i+ \frac{1}{f}\sum\limits_{k=1}^{i-1} \frac{f-(i-k)}{f}(\mathcal{X}_c^{i}-\mathcal{X}_c^{k}), \ \ \forall i = 2, \ldots, f
\label{eq:ssa_layer2}
\end{equation}
Here, $k$ is the shift and $i$ is a location along the temporal direction. If $i= 1$, $\mathcal{X}_s^i =\mathcal{X}_c^i$. Since, the nearby frames can have more contextual relation, the term $\frac{f-(i-k)}{f}$ is to ensure that the larger shifts get smaller weights than smaller shifts.\\
Instead of computing for each temporal volume separately, $X_s$ can be computed in a cumulative manner as illustrated in Fig. \ref{fig:ssa_layer}. However, the mathematical formulation and the illustration shown in Fig. \ref{fig:ssa_layer} lead to the same output. Since, $\mathcal{X}_c$ is a four dimensional volume, it would be hard to provide a clean illustration. We have also omitted the multiplicative constants from the illustration to keep it clean. Hence, we have used 1-D representation to illustrate its operations visually. In Fig. \ref{fig:ssa_layer}, each column refers to a single shift. It can be seen that the input feature map $\mathcal{X}_c$ is subtracted from its shifted version and then, the difference is added to it in the corresponding locations to obtain $\hat{\mathcal{X}_s}$. Then, we again shift the feature map $\mathcal{X}_c$ by one more step, subtract it from its original version and then add the difference to the corresponding locations of $\hat{\mathcal{X}_s}$. At the end of $f-1$ steps, we obtain $\mathcal{X}_s$.\\
\textbf{Temporal pooling.} As, we move along the depth of the 3D convolution networks, the temporal depth of the feature maps keeps reducing as we perform 3D convolutions of stride more than one. In our case, we are not performing convolution along the temporal depth. Hence, to reduce the temporal depth, we perform max pooling along the temporal direction whenever we want to reduce the temporal depth of the feature maps.\\
\textbf{Parameter Analysis.} A standard 3D convolutional kernel of size $n\times n\times n$ and $c$ channels contains $cn^3$ parameters. The proposed convolution block comprises a standard 2D-convolution kernel, an SSA layer and temporal max pooling. A standard 2D-convolution kernel of size $n\times n$ and $c$ channels contains $cn^2$ parameters and an SSA layer consists of shift, subtract and add operations which do not require any trainable parameters. Also, temporal max pooling does not require any trainable parameters. Hence, the overall number of trainable parameters used in the proposed convolution block is $cn^2$ which is $n$ times less than the standard 3D convolution kernel.
\section{Experiments and Discussions}
\label{sec:experiments}
\begin{figure}[t]
	\centering
	\begin{minipage}{\linewidth}
		\centering
		\stackunder{\includegraphics[width=0.15\linewidth]{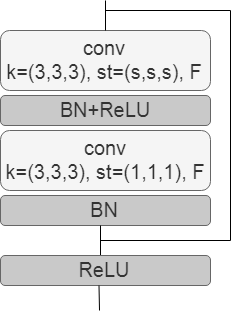}}{\small (a) ResNet (Basic)}\hspace{2em}
		\stackunder{\includegraphics[width=0.15\linewidth]{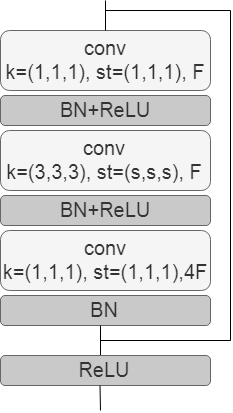}}{\small (b) ResNet (Bottleneck)}\hspace{2em}
		\stackunder{\includegraphics[width=0.2\linewidth]{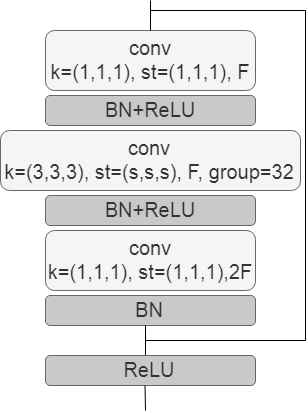}}{\small (c) ResNeXt (Bottleneck)}\hspace{2em}
	\end{minipage}
	\begin{minipage}{\linewidth}
		\centering
		\stackunder{\includegraphics[width=0.2\linewidth]{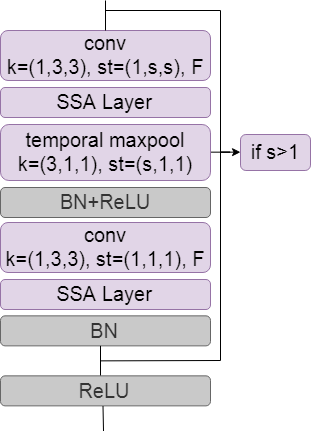}}{\small (d) SSA-ResNet (Basic)}
		\stackunder{\includegraphics[width=0.2\linewidth]{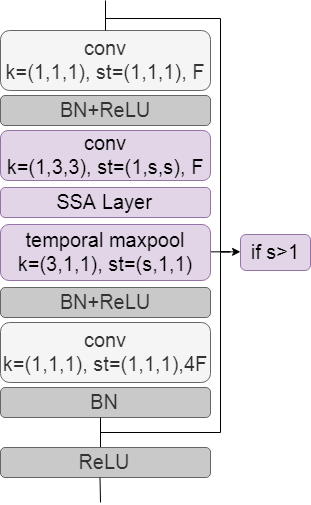}}{\small (e) SSA-ResNet (Bottleneck)}
		\stackunder{\includegraphics[width=0.25\linewidth]{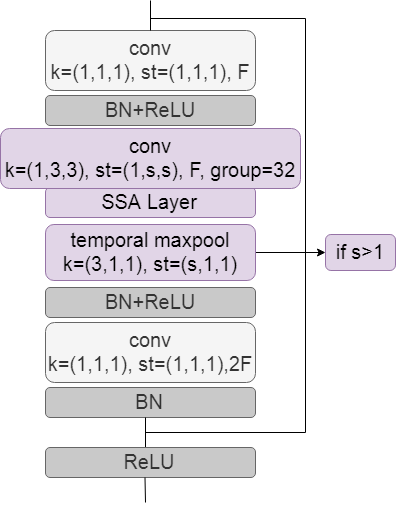}}{\small (f) SSA-ResNeXt (Bottleneck)}
	\end{minipage}
	\caption{(a) and (b) show the basic and bottleneck blocks used in 3D ResNet architecture \cite{hara2018can}. (c) shows the bottleneck bock used in 3D ResNeXT architecture \cite{hara2018can}. (d), (e) and (f) show the residual blocks in which 3D convolution kernel is replaced by the proposed convolutional block.}\vspace{-1.5em}
	\label{fig:blocks}
\end{figure}
In this section, we show that the 3D CNNs perform better when the standard 3D convolution kernels are replaced by the proposed convolutional block. Our focus is mostly on the residual networks. We evaluate their performances on two types of 3D data: spatio-temporal image sequences and 3D CAD models.
\subsection{Spatio-Temporal Image Sequences}
\textbf{Dataset.} UCF101 \cite{soomro2012ucf101} is a benchmark action recognition dataset containing complex real world videos which has been used in several works\cite{tran2015learning,tran2017convnet,diba2018spatio}. The videos of the dataset cover 101 action categories. We use UCF101 split-1 for all our experiments regarding spatio-temporal image sequences.
\subsubsection{Network Architectures}
\label{sec:network_arch}
We employ deep 3D residual networks to evaluate the proposed convolutional block \cite{hara2018can}. Fig. \ref{fig:blocks} (a) and (b) show the basic and bottleneck block used in 3D ResNet architecture \cite{hara2018can}. Fig. \ref{fig:blocks} (c) shows the bottleneck block used in ResNeXT architecture \cite{hara2018can}. We replace the 3D convolution kernel of the residual blocks by the proposed convolutional block as shown in Fig. \ref{fig:blocks} (d), (e) and (f). In Fig. \ref{fig:blocks} (d), (e) and (f), it can be seen that we preserve the overall structure of the blocks while replacing the 3D convolution kernel by the proposed convolutional block. This is done to show the true effect of the proposed block on the existing networks. We have also experimented with the WideResNet architecture with a widening factor of 2 \cite{hara2018can}. The structure of bottleneck block of WideResNet is same as the bottleneck block of ResNet. The only difference is the number of channels of the feature maps in the layers. To show that the proposed approach is not constrained to the residual networks, we have also done experiments with C3D network proposed in \cite{tran2015learning}. Similar to the residual networks, we replace the 3D convolution kernel with the proposed convolutional block.  The training details are provided in the supplementary material.
\begin{table}[t]
	\centering \scriptsize
	\begin{tabular}{|c|c|c|c|c|c|}
		\hline
		Network & Layers & \makecell{Parameters\\(Millions)} & \makecell{SSA Layer} & \makecell{Temporal pooling} & Accuracy(\%)  \\ \hline
		3D ResNet\cite{hara2018can,tran2017convnet} (baseline) & 18 &  $\approx$ 33 &  & & 45.6   \\
		SSA-ResNet (ours) & 18 & $\approx$ 11  &  & \checkmark& 52.8 \\
		SSA-ResNet (ours) & 18 & $\approx$ 11  & \checkmark & \checkmark& \textbf{55.7} \\ \hline
		3D ResNeXT\cite{hara2018can} (baseline) & 50 & $\approx$ 26  & & & 49.3    \\
		SSA-ResNeXT (ours) & 50 & $\approx$ 23 & &\checkmark & 54.9  \\
		SSA-ResNeXT (ours) & 50 & $\approx$ 23 & \checkmark& \checkmark&  \textbf{56.9} \\\hline
		3D WideResNet\cite{hara2018can} (baseline) & 50 & $\approx$ 157 &  & & 46.8   \\
		SSA-WideResNet(ours) & 50 & $\approx$ 67 & & \checkmark& 50.7  \\
		SSA-WideResNet(ours) & 50 & $\approx$ 67 & \checkmark& \checkmark& \textbf{52.9}  \\\hline
		C3D\cite{tran2015learning} (baseline) &5  & $\approx$ 18 &  & &     44  \\
		SSA-C3D (ours) &5  & $\approx$ 14  &     & \checkmark&   50 \\
		SSA-C3D (ours) &5  & $\approx$ 14  &     \checkmark & \checkmark&  \textbf{51.6}  \\\hline
		3D ResNet\cite{diba2018spatio,hara2018can} (baseline) & 101 &  $\approx$ 88 &  & &  46.7  \\
		SSA-ResNet (ours) & 101 & $\approx$ 43  &  & \checkmark& 52.1 \\
		SSA-ResNet (ours) & 101 & $\approx$ 43  & \checkmark & \checkmark& \textbf{54.4} \\ \hline
	\end{tabular}
	\caption{\textbf{Comparisons with baselines.} The comparison of the test accuracies obtained by the baseline 3D models with the networks obtained by replacing the 3D convolution kernel by the proposed convolution block in the baseline 3D models on UCF101 split-1 when trained from scratch.}\vspace{-1.5em}
	\label{tab:ablation_baseline}
\end{table}
\subsubsection{Comparisons with baselines}
We perform our experiments by training the networks from scratch on UCF101 split-1. The test accuracies of 3D ResNeXT and 3D WideResNet when trained from scratch on UCF101 split-1 are not available in the previous works\cite{hara2018can}. So, we train these networks on UCF101 from scratch to obtain them. For the other baseline networks, we mention the accuracies reported in \cite{hara2018can,diba2018spatio,tran2017convnet}. SSA-ResNet, SSA-WideResNet, SSA-ResNeXT, and SSA-C3D are obtained by replacing the 3D convolution kernels in ResNet, WideResNet, ResNeXT, and C3D\cite{tran2015learning} by the proposed convolutional block. We train these networks from scratch on UCF101 with same hyperparameter settings. Table \ref{tab:ablation_baseline} shows that the accuracies obtained by the baseline 3D models when trained from scratch on the split-1 of UCF101 dataset. It also shows the accuracies obtained by replacing the 3D convolution kernel by the proposed convolution block. It can be seen that the networks perform significantly better with the proposed convolutional block while utilizing lesser trainable parameters.
\begin{table}[t]
	\centering  \scriptsize
	\begin{tabular}{|c|c|c|c|c|}
		\hline
		Network& Layers & \makecell{Parameters (Millions)} & \makecell{Model Size (MB)} & Accuracy  \\ \hline
		2D-ResNet\cite{he2016deep,tran2017convnet} &  18 & $\approx$11.2 & - &  42.2 \\
		2D-ResNet\cite{he2016deep,tran2017convnet} &  34 & $\approx$21.5 & - &  42.2 \\
		3D-ResNet\cite{tran2017convnet} &  18 & $\approx$33.2 & 254 & 45.6 \\
		3D-ResNet\cite{tran2017convnet} &  34 & $\approx$63.5 & 485 & 45.9 \\
		3D-ResNet\cite{diba2018spatio} &  101 & $\approx$86.06 & 657 & 46.7 \\
		3D STC-ResNet\cite{diba2018spatio} &  18 & - & - & 42.8 \\
		3D STC-ResNet\cite{diba2018spatio} &  50 & - & - &46.2 \\
		3D STC-ResNet\cite{diba2018spatio} &  101 & - & - &47.9 \\
		C3D\cite{tran2015learning} &  5 & $\approx$18 & 139.6 &44 \\
		R(2+1)D\cite{tran2018closer} &  18 & $\approx$33.3 & 128 &48.37\\\hline
		SSA-ResNet (ours) &  18 & $\approx$11 & 88.5 &55.7 \\
		SSA-ResNeXt (ours) &  50 & $\approx$23 & 185.9 &\textbf{56.9}
		\\\hline
	\end{tabular}
	\caption{\textbf{Comparisons with the state-of-the-art. } The comparison of the proposed approach with the state-of-the-art methods when trained from scratch on UCF101 dataset.}\vspace{-1.5em}
	\label{tab:results}
\end{table}
\subsubsection{Comparisons with the state-of-the-art}
Table \ref{tab:results} compares the proposed approach with the state-of-the-art methods when trained from scratch on UCF101 dataset. The test accuracy of R(2+1)D\cite{tran2018closer} when trained from scratch on UCF101 is not available in the previous works. So, we trained the network on UCF101 split-1 from scratch to obtain it using the same hyperparameter settings as ours. It can be observed that SSA-ResNeXT performs significantly better than the previous approaches. SSA-ResNet (18 layers) utilizes approximately 11 million parameters which is roughly equal to the parameters used in 2D-ResNet \cite{he2016deep} (18 layers). Inspite of having almost equal parameters, SSA-ResNet (18 layers) outperforms 2D-ResNet (18 layers) by 13.5 \% in terms of classification accuracy. Also, SSA-ResNet (18 layers) utilizes approximately 3 times less parameters than 3D-ResNet (18 layers)\cite{tran2017convnet}, 3D STC-ResNet (18 layers)\cite{diba2018spatio}, and R(2+1)D (18 layers)\cite{tran2018closer} and still outperforms them by 10.1\%, 12.9 \%, and 7.33\%, respectively.
\begin{table}[t]
	\centering \scriptsize
	\begin{tabular}{|c|c|c|}
		\hline
		\#Shift & Temporal pooling & Accuracy \\\hline
		0 & & 46.3 \\
		0 & \checkmark & 52.8\\
		1 & \checkmark & 52.6 \\
		2 & \checkmark & 53.4 \\
		3 & \checkmark & 53.9 \\
		f-1 &  & 51.3 \\
		f-1 & \checkmark & \textbf{55.7}
		\\\hline
	\end{tabular}
	\caption{\textbf{Analysis of different shifts and temporal pooling}. The comparison of test accuracies obtained on UCF101 split-1 using SSA-ResNet (18 layers) (when trained-from-scratch) with varying number of shifts along with the effect of temporal pooling.}\vspace{-2em}
	\label{tab:ablation_shift}
\end{table}
\vspace{-0.5em}
\subsubsection{Analysis}
In the proposed convolutional block, apart from a standard 2D convolution kernel, there are two components: SSA layer and Temporal pooling.\\
\textbf{SSA Layer.} As shown in Fig.\ref{fig:ssa_layer}, we perform the shift operation $f-1$ times, where $f$ is the temporal depth of the input feature map. We perform the experiments on SSA-ResNet (18 layers) with different values of shifts. The results are shown in Table \ref{tab:ablation_shift}. It can be seen that as we increase the fixed number of shifts from 1 to 3, the test accuracy increases and we obtain the highest accuracy when we perform $f-1$ shifts.\\
\textbf{Temporal Pooling.} In Table \ref{tab:ablation_shift}, it can be observed that by using 2D-convolution kernel and only max temporal pooling, the network outperforms the baseline case, i.e. with only 2D convolution kernels. The same pattern can be observed in Table \ref{tab:ablation_baseline}, in which the baseline 3D models are replaced with the proposed convolution block without SSA layer (second row for each network) and the networks performed significantly better than the baseline 3D CNNs.
\begin{table}[t]
	\centering \scriptsize
	\begin{tabular}{|c|c|c|c|c|c|} \hline
		Network & Framework & Augmentation & \makecell{Parameters\\(Millions)} & ModelNet40 (\%) & ModelNet10 (\%)  \\\hline
		3D ShapeNets\cite{wu20153d} &  Volumetric & Az $\times$ 12 & $\approx$38  & 77  & 83.5 \\
		Beam Search\cite{xu2016beam} &  Volumetric & Az $\times$ 12 & $\approx$0.08  & 81.26 & 88 \\
		3D-GAN\cite{wu2016learning} &  Volumetric & Az $\times$ 12 & $\approx$11 & 83.3 & 91 \\
		VoxNet\cite{maturana2015voxnet} &  Volumetric & Az $\times$ 12 & $\approx$0.92 & 83 & 92 \\
		LightNet\cite{zhi2017lightnet} &  Volumetric & Az $\times$ 12 & $\approx$0.30 & 86.90 & 93.39\\
		ORION\cite{sedaghat2017orientation} &  Volumetric & Az $\times$ 12 & $\approx$.91 & - & \textbf{93.8} \\\hline
		SSA-ResNeXT8 (ours) &  Volumetric & Az $\times$ 12 & $\approx$3.38 & \textbf{89.5} & 93.3 \\ \hline
	\end{tabular}
	\caption{\textbf{Comparisons with the state-of-the-art.} The comparison of the SSA-ResNeXT8 with the state-of-the-art methods on the voxelized version of ModelNet40 and ModelNet10 datasets.}\vspace{-1em}
	\label{tab:modelnet}
\end{table}
\subsection{3D CAD Models}
\textbf{Dataset.} ModelNet\cite{wu20153d} is a collection of 3D CAD models of objects. It has two subsets: ModelNet10 and ModelNet40. ModelNet10 and ModelNet40 contains 10 and 40 classes of objects, respectively, which are manually aligned to a canonical frame. In our experiments, we use the voxelized version of size $32\times 32 \times 32$ and augmentation with 12 orientations \cite{maturana2015voxnet}. Similar to \cite{maturana2015voxnet,brock2016generative}, we add noise, random translations, and horizontal flips for data augmentation to the training data. Similar to \cite{brock2016generative}, we scale the binary voxel range from $\{0,1\}$ to $\{-1,5\}$.\\
\textbf{Network Architecture. } 
To avoid overfitting on ModelNet40 and ModelNet10, we use a smal network SSA-ResNext8 to evaluate our approach on 3D CAD models. We use the SSA-ResNeXT bottleneck block in the architecture of the network. Let us denote the SSA-ResNeXT bottleneck block with $SSAR(k,F,s)$, where $1\times k\times k$ is the kernel size of the 2D convolution filter, $F$ is the number of channels in the input feature map and  $s$ is the value of stride passed to the block. The architecture of SSA-ResNext8  is as follows: $Conv2D(3,1)\rightarrow MP(3,2)\rightarrow SSAR(3,64,1)\rightarrow SSAR(3,256,1)\rightarrow SSAR(3,256,2)\rightarrow SSAR(3,512,1)\rightarrow SSAR(3,512,2)\rightarrow SSAR(3,1024,1)$ $\rightarrow GP\rightarrow FC$. Here, $Conv2D(3,1)$ is a 2D convolution kernel of size $1\times 3\times 3$ and stride of 1, followed by a batch normalization layer and ReLU, and $MP(3,2)$ is the max-pooling layer with kernel size of $3\times 3\times 3$ and stride of 2. $GP$ and $FC$ stands for global average pooling and fully connected layer, respectively. The training details are provided in the supplementary material.\\
\textbf{Comparisons with the state of the art.} Table \ref{tab:modelnet} shows the comparison of the SSA-ResNeXT8 with the state-of-the-art methods that use voxelized/volumetric ModelNet datasets as input. For fair comparison, we only consider volumetric frameworks. It can be observed that the network with the proposed convolutional block performs better than the state-of-the-art on ModelNet40 and comparable on ModelNet10 in the case when the networks are trained with shapes augmented with 12 orientations. This shows that the proposed convolution block is not restricted to videos and can be further exploited in shapes.
\section{Conclusion}
We propose a novel convolutional block which is proficient in capturing both spatial and temporal structure  of the 3D data while utilizing lesser parameters than the 3D convolution kernel. It comprises three components: a 2D-convolution kernel to capture the spatial information, a novel SSA layer to capture the temporal structure, and a temporal pooling layer to reduce the temporal depth of the input feature map. We show that the 3D CNNs perform better when the 3D convolution kernels are replaced by the proposed convolutional block. SSA-ResNet (18 layers) outperforms the state-of-the-art accuracy on the UCF101 dataset split-1 while utilizing lesser parameters when networks are trained-from-scratch. We have also evaluated the proposed convolutional block on 3D CAD models and we outperform the state-of-the-art on ModelNet40 among the volumetric framework, when the training data is augmented with 12 rotations.
%
%
%
%

\end{document}